\documentclass[12pt]{article}

\let\citep\cite
\let\citet\cite

\usepackage[utf8]{inputenc}
\usepackage[T1]{fontenc}
\usepackage{amsmath}
\usepackage{graphicx}
\usepackage{hyperref}
\usepackage{cleveref} 
\usepackage{multirow}
\usepackage{adjustbox}
\usepackage{bm}
\usepackage{etoolbox}
\usepackage{booktabs}
\usepackage{array}
\usepackage{algorithm}
\usepackage[noend]{algorithmic}
\usepackage{setspace}
\usepackage{authblk}
\usepackage{graphicx}     
\usepackage{booktabs}     
\usepackage{multirow}     
\usepackage{pifont}       
\usepackage{amssymb}      

\usepackage{pifont}

\newcommand{\cmark}{\checkmark}
\newcommand{\xmark}{\ding{55}}

\newcommand{\Description}[1]{}

\providecommand{\keywords}[1]{\textbf{Keywords:} #1}

\begin{document}

\title{SHAPoint: Task‑Agnostic, Efficient, and Interpretable Point‑Based Risk Scoring via Shapley Values}


\title{SHAPoint: Task‑Agnostic, Efficient, and Interpretable Point‑Based Risk Scoring via Shapley Values}
\author[1]{Tomer D.\ Meirman}
\author[1]{Bracha Shapira}
\author[1,2]{Noa Dagan}
\author[1]{Lior S.\ Rokach}
\affil[1]{Faculty of Computer and Information Sciences, Ben-Gurion University of the Negev, Beer–Sheva, Israel}
\affil[2]{Clalit Research Institute, Clalit Health Services, Tel‑Aviv, Israel}
\date{\today} 

\maketitle
\begin{abstract}
Interpretable risk scores play a vital role in clinical decision support, yet traditional methods for deriving such scores often rely on manual preprocessing, task-specific modeling, and simplified assumptions that limit their flexibility and predictive power. We present SHAPoint, a novel, task-agnostic framework that integrates the predictive accuracy of gradient boosted trees with the interpretability of point-based risk scores. SHAPoint supports classification, regression, and survival tasks, while also inheriting valuable properties from tree-based models, such as native handling of missing data and support for monotonic constraints. Compared to existing frameworks, SHAPoint offers superior flexibility, reduced reliance on manual preprocessing, and faster runtime performance. Empirical results show that SHAPoint produces compact and interpretable scores with predictive performance comparable to state-of-the-art methods, but at a fraction of the runtime, making it a powerful tool for transparent and scalable risk stratification.

\end{abstract}




\section{Introduction}

\begin{figure}[htbp]
    \centering
    \includegraphics[width=\textwidth]{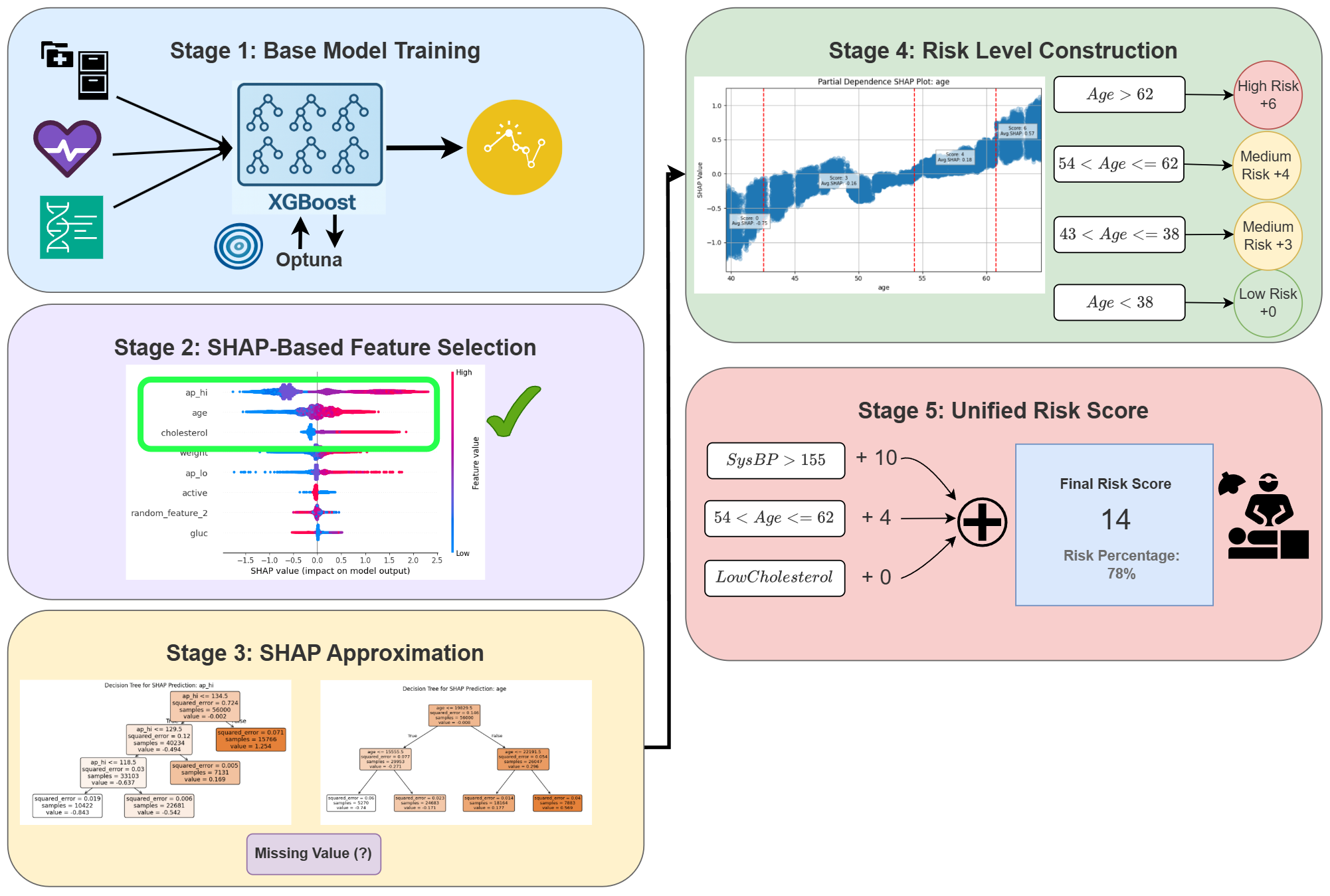}
    \Description{Graphical abstract summarizing the five‑stage SHAPoint pipeline.}
    \caption{The SHAPoint Method for Interpretable Risk Scoring. This five-stage pipeline integrates machine learning accuracy with clinical interpretability. \textbf{Stage 1: Base Model Training} uses an ensemble model (e.g., XGBoost) with hyperparameter optimization (e.g., Optuna) to learn from input data. \textbf{Stage 2: SHAP-Based Feature Selection} identifies the most influential features by quantifying their contributions to model predictions. \textbf{Stage 3: SHAP Approximation} transforms complex SHAP values into simple, univariate decision trees, partitioning the feature space into distinct, interpretable bins. \textbf{Stage 4: Risk Level Construction} converts these tree-based rules into intuitive risk levels, assigning scores based on feature contributions and inherently handling missing values. Finally, \textbf{Stage 5: Unified Risk Score} aggregates individual risk levels into a single, transparent, and scaled risk score, suitable for various predictive tasks.}
    \label{fig:graphical_abstract}
\end{figure}

Point-based clinical risk scores are indispensable tools in modern decision support systems \citep{steyerberg2009clinical}. From triage in emergency departments to long-term outcome prediction in chronic care, risk scores provide a compact, interpretable summary of patient risk based on a small set of variables. Traditionally, such scores have been derived from logistic regression models using a fixed number of manually selected variables, discretized into coarse categories, and assigned integer weights \citep{wilson1998prediction, antman2000timi}. While this approach supports ease of use and clinical interpretability, it often suffers from reduced predictive performance, rigid assumptions, and limited adaptability across tasks or domains, with the optimal handling of continuous predictors remaining a key challenge \citep{molero2024optimal}. 

Recent advances in machine learning, especially ensemble-based models like gradient boosted trees (e.g., XGBoost, CatBoost) \citep{chen2016xgboost, dorogush2018catboost}, have dramatically improved predictive accuracy in healthcare applications. However, these models are typically treated as "black boxes"—while they perform well, their outputs are difficult to interpret and rarely translated into clinical scoring systems. Bridging the gap between model performance and interpretability has emerged as a major challenge in predictive modeling, particularly in high-stakes domains such as healthcare, finance, and public policy \citep{rudin2019stop, ribeiro2016should, molnar2020interpretable}. Recent literature in explainable AI (XAI) has moved beyond simply g
enerating explanations to critically evaluating their faithfulness to the underlying model and their utility in high-stakes human decision-making, demanding methods that ensure transparency, accountability, and trust \cite{chen2022interpretable, mersha2024explainable}.

Various frameworks have been proposed that aim to combine machine learning with interpretable scoring. AutoScore\citep{xie2020autoscore, Xie2022AutoScoreSurvival}, for instance, is a widely-used pipeline that was recently extended into a universal framework for generating scores for binary, survival, and ordinal outcomes. It typically employs a multi-stage process, first using a machine learning model for variable ranking before deriving final point values through a separate, simpler model like logistic or Cox regression. Other approaches focus on the robustness of explanations. FasterRisk\cite{liu2022fasterrisk}, introduces a beam search algorithm combined with star ray search for efficient risk score generation, producing a pool of almost-optimal sparse continuous solutions that are transformed into integer-coefficient risk scores. The Shapley Variable Importance Cloud \citep{ning2022shapley} (ShapleyVIC), for example, provides a statistically principled method to assess whether a variable's importance is stable across an entire set of high-performing models (the "Rashomon set"), thereby guarding against over-claiming importance based on a single model. Other novel approaches include SymScore, which leverages symbolic regression to generate transparent scoring models directly from data \citep{cawiding2025symscore}. While these methods offer valuable innovations, they often address different goals and carry distinct limitations. Frameworks like ShapleyVIC, designed for robustness analysis, are computationally intensive by nature, while AutoScore's multi-stage process can create a disconnect between the model used for variable selection and the one used for final score derivation. This can lead to scores that are not fully aligned with the complex interactions captured by the initial machine learning model.

In this paper, we propose SHAPoint, a novel framework that unifies the strengths of model explainability, machine learning, and interpretable scoring into a single, task-agnostic pipeline. SHAPoint is built on three core principles: efficiency, interpretability, and flexibility. In contrast to previous methods, SHAPoint uniquely combines model-agnostic flexibility, automated binning, native handling of missing values, and computational efficiency into a unified scoring system. 

First, the method begins by training a high-performing gradient boosted trees (GBT) model (e.g., XGBoost) on the full feature set, regardless of whether the prediction task involves classification, regression, or survival analysis. This ensures that the base model captures complex, non-linear interactions without sacrificing accuracy. Second, SHAP \citep{lundberg2017unified} (SHapley Additive exPlanations) values are computed to quantify each feature’s contribution to the model’s predictions. Top-K important variables are then selected based on their SHAP values, providing a robust, model-aligned measure of variable relevance.

The third and most distinctive step in SHAPoint involves transforming each selected feature into interpretable, score-generating levels using a decision tree trained to approximate the feature’s SHAP contribution. This tree, constrained to a fixed number of leaves, partitions the feature space into distinct risk levels. The leaf with the lowest average SHAP value is set as the reference level with a score of zero, and other levels receive scaled point values corresponding to their relative contribution. This process converts continuous or complex features into compact, clinically meaningful categories without relying on manual binning or arbitrary thresholds.

After repeating this process for all selected variables, the model assembles a comprehensive risk scoring table by unifying all feature-level point values. Scores are normalized and scaled into small, easily interpretable integers. The final product is a fully interpretable risk model that maintains a close alignment with the high-performing underlying machine learning model. Importantly, SHAPoint inherits the task-agnostic nature of SHAP \citep{lundberg2017unified} and GBTs (such as XGBoost \citep{chen2016xgboost} and CatBoost \citep{dorogush2018catboost}), making it suitable for a wide range of prediction tasks—including binary classification (e.g., sepsis prediction), continuous regression (e.g., hospital length of stay), and time-to-event survival analysis (e.g., mortality risk) \citep{guo2023factor, tarabanis2023explainable, jo2021prediction}.

The advantages of SHAPoint are numerous. It offers model-agnostic flexibility, automated binning, native handling of missing values, and computational efficiency, overcoming limitations of existing methods like AutoScore and ShapleyVIC. To evaluate the proposed method, we conducted a comprehensive comparative study against state-of-the-art baselines such as FasterRisk, ShapleyVIC, and AutoScore-Survival, using diverse publicly available healthcare datasets including Cardiovascular Disease, Breast Cancer, and MIMIC-III ICU. Our evaluation emphasizes predictive performance (ROC AUC, PR AUC, C-index) and computational efficiency.

\section{Methods}
We propose SHAPoint, a point-based risk scoring method that combines the predictive accuracy of ensemble methods with the interpretability requirements of risk assessment applications. Our approach leverages Shapley Additive Explanations (SHAP \citep{lundberg2017unified}) values to identify the most influential features and approximates their contributions using decision tree-based rules, resulting in an interpretable scoring system with transparent decision boundaries.
The proposed framework addresses the fundamental trade-off between model accuracy and interpretability by decomposing the prediction process into five sequential steps: (1) training a high-accuracy base model, (2) identifying the most influential features using SHAP values, (3) approximating individual feature SHAP contributions with decision trees and using the decision tree structure to bin the features, (4) converting tree structures into interpretable risk levels, and (5) unifying all levels into a coherent scoring system. This approach ensures that the final model maintains predictive performance while providing clear, actionable insights for risk assessment practitioners.

\subsection{Base Model Training}
Given a dataset $\mathcal{D} = \{(x_i, y_i)\}_{i=1}^n$ with $x_i \in \mathbb{R}^p$ and $y_i \in \mathbb{R}$, Given a dataset... we train a gradient boosted tree model (e.g., XGBoost, CatBoost, or LightGBM) to capture nonlinear patterns. The model is trained to minimize the regularized objective:

\[
\mathcal{L} = \sum_{i=1}^n l(y_i, \hat{y}_i) + \sum_{k=1}^K \Omega(f_k)
\]

where $\hat{y}_i = \sum_{k=1}^K f_k(x_i)$ is the predicted output, $f_k$ denotes the $k$-th tree, and $\Omega(f_k)$ is a regularization term. The choice of loss function depends on the nature of the learning task and, by default, leverages built-in objective functions in XGBoost \citep{chen2016xgboost} (similar objective functions can be find in other GBTs such as CatBoost \cite{dorogush2018catboost}). Specifically, binary:logistic is used for binary classification tasks, reg:squarederror for regression tasks, and survival:cox for survival analysis.

Other hyperparameters that can be set by the user include:
\begin{itemize}
    \item \texttt{n\_estimators}: Number of boosting rounds (100–1000),
    \item \texttt{max\_depth}: Tree depth (3–10),
    \item \texttt{learning\_rate}: Step size (0.01–0.3),
    \item \texttt{subsample}: Sample fraction per tree (0.8–1.0),
    \item \texttt{colsample\_bytree}: Feature fraction per tree (0.8–1.0).
\end{itemize}

These hyperparameters can be configured manually or optimized automatically using Bayesian optimization. In our implementation, we employ the Optuna \cite{akiba2019optuna} framework to efficiently search for the optimal hyperparameter configuration.  

\subsection{Feature Selection via SHAP Analysis}

SHAP values decompose the model prediction into contributions from each feature:

\[
f(x) = \phi_0 + \sum_{j=1}^p \phi_j(x)
\]

The global importance of feature $j$ is computed as the mean absolute SHAP value:

\[
I_j = \frac{1}{n} \sum_{i=1}^n |\phi_j(x_i)|
\]

We select the top $K$ features using $I_j$, halting early if a randomly injected feature appears in the list. The user can determine how many (if any) random features will be augmented into the original dataset. This approach ensures robustness by filtering out noise and preserving interpretability \citep{ning2022shapley}.

\subsection{SHAP Value Approximation with Decision Trees}
We aim to use SHAP values to quantify the contribution of each feature in the final scoring model. A distinctive feature of SHAPoint is its ability to determine optimal binning of each input feature based on the distribution of SHAP values across the entire model, rather than directly from the target variable. 

Predicting SHAP values rather than binning directly on the target allows the method to account for the complex ways and potentially non-linear ways features influence risk in the context of the full model, resulting in bins that are both more interpretable and more aligned with the actual decision logic of the model. This approach enhances both the transparency and the fidelity of the risk scores, ensuring that each bin reflects a meaningful contribution to model output rather than just a marginal association with the outcome.

To create interpretable partitions, we train univariate regression trees $T_j$ for each selected feature $j$ using the SHAP values:

\[
T_j(x_{i,j}) \approx \phi_j(x_i)
\]

The trees are constrained by either a maximum leaf count $M$ with $\text{max\_depth} = \lfloor \log_2 M \rfloor + 1$, or user-defined $max\_depth$. 

Despite the setting by the user, we apply Minimal Cost-Complexity Pruning with cross-validation to prevent overfitting \cite{breiman1984cart}. This method balances model complexity and training error by introducing a penalty term for the number of terminal nodes in the tree, allowing us to prune branches that do not contribute significantly to predictive performance. By selecting the optimal complexity parameter via cross-validation, we ensure that the final model generalizes well to unseen data, and that the final scoring model will not include more levels than needed.

\subsection{Risk Level Construction}

Each leaf $\ell \in T_j$ represents a feature bin with average SHAP value $s_\ell$. Risk levels are computed relative to the lowest-risk leaf used as a reference value with zero score:

\[
\text{score}_\ell = s_\ell - \min_{\ell' \in T_j} s_{\ell'}
\]

Rules for each bin are extracted from decision paths, e.g.:

\[
R_\ell: x_j \in [a_\ell, b_\ell] \quad \text{or} \quad (-\infty, a_\ell], \quad (b_\ell, \infty)
\]

Every path from the root to a leaf node can be translated into a decision rule represented as a conjunction of literals (e.g., conditions such as $a < 10$ and $b \geq 5$). However, some predicates along the path may be redundant. For example, if a path includes both $a < 10$ and $a < 8$, the condition $a < 10$ becomes superfluous, as it is already implied by $a < 8$. To enhance interpretability and reduce cognitive load, we simplify these rules by eliminating such redundant constraints, resulting in more concise and human-readable decision rules without altering their predictive meaning \citep{IoTaPlus_OSRE}.
 
To address missing values in the input variables, our methodology eschews standard imputation techniques in favor of a data-driven approach that is integrated directly into the risk scoring framework. We leverage the inherent capability of decision trees to handle missing data during the training and prediction phase \cite{breiman1984cart}. When a new observation with a missing value is scored, the corresponding decision tree model assigns it to a specific leaf node based on its internal learned structure. This process results in the missing value being assigned a specific, empirically-derived risk score associated with that leaf's rule. This technique allows the model to quantify the risk associated with the absence of information for a given variable, thereby incorporating it as a meaningful factor in the final aggregated risk score.

\subsection{Unified Risk Score Construction}

To generate an integrated risk score, we scale all risk levels to integer values in $[0, S_{\max}]$:

\[
\text{scaled\_score}_\ell = \left\lfloor S_{\max} \cdot \frac{\text{score}_\ell - \min_{\ell' \in \mathcal{L}} \text{score}_{\ell'}}{\max_{\ell' \in \mathcal{L}} \text{score}_{\ell'} - \min_{\ell' \in \mathcal{L}} \text{score}_{\ell'}} \right\rfloor
\]

where $\mathcal{L} = \bigcup_{j=1}^K \mathcal{L}_j$ is the set of all leaf bins. For a new instance $x^*$, the total risk score is:

\[
\text{RiskScore}(x^*) = \sum_{j=1}^K \text{scaled\_score}_{\ell_j^*}
\]

where $\ell_j^*$ is the bin that $x_j^*$ falls into.

\subsection{Algorithm Summary}

\Cref{alg:shaprisk} summarizes the full pipeline, which supports classification, regression, and survival tasks with minimal preprocessing. The model ensures high fidelity to the underlying XGBoost predictions while remaining interpretable and auditable.

{\small
\begin{algorithm*}[!t]
\caption{SHAP-Based Interpretable Risk Scoring}
\label{alg:shaprisk}
\begin{algorithmic}[1]
\setstretch{0.45}
\REQUIRE Training data $\mathcal{D} = \{(x_i, y_i)\}_{i=1}^n$, Hyperparameters: $K$, $M$, $S_{\max}$, \texttt{base\_model\_params}, \texttt{use\_optuna}
\ENSURE Unified risk scoring model

\STATE \textbf{1. Train Base Model}
\STATE \texttt{model} $\leftarrow$ \texttt{XGBoost}$(\mathcal{D}, \texttt{base\_model\_params}, \texttt{use\_optuna})$

\STATE \textbf{2. Feature Selection}
\STATE $\Phi \leftarrow$ \texttt{SHAP\_values}(\texttt{model}, $X$)
\STATE \texttt{importance} $\leftarrow \texttt{mean}(|\Phi|, \texttt{axis}=0)$
\STATE \texttt{top\_features} $\leftarrow \texttt{select\_top\_K}(\texttt{importance}, K)$

\STATE \textbf{3. Train SHAP Approximation Trees}
\FOR{each feature $j$ in \texttt{top\_features}}
    \STATE \texttt{max\_depth} $\leftarrow \lfloor \log_2 M + 1\rfloor$
    \STATE $T_j \leftarrow \texttt{DecisionTree}(X_j, \Phi_j, \texttt{max\_depth})$
    \STATE \texttt{shap\_trees}[$j$] $\leftarrow T_j$
\ENDFOR

\STATE \textbf{4. Create Risk Levels}
\STATE \texttt{all\_levels} $\leftarrow$ [ ]
\FOR{each feature $j$ in \texttt{top\_features}}
    \STATE \texttt{leaves} $\leftarrow \texttt{extract\_leaves}(T_j)$
    \STATE \texttt{rules} $\leftarrow \texttt{extract\_rules}(T_j, j)$
    \STATE \texttt{min\_score} $\leftarrow \min(\texttt{leaf\_predictions}(T_j))$
    \FOR{each leaf $\ell$ in \texttt{leaves}}
        \STATE \texttt{score} $\leftarrow \texttt{leaf\_prediction}(\ell) - \texttt{min\_score}$
        \STATE \texttt{rule} $\leftarrow \texttt{simplify\_rule}(\texttt{rules}[\ell])$
        \STATE \texttt{all\_levels.append(\{feature: $j$, leaf: $\ell$, score: score, rule: rule\})}
    \ENDFOR
\ENDFOR

\STATE \textbf{5. Unify Scoring System}
\STATE \texttt{min\_global} $\leftarrow \min(\texttt{level.score for level in all\_levels})$
\STATE \texttt{max\_global} $\leftarrow \max(\texttt{level.score for level in all\_levels})$
\FOR{each \texttt{level} in \texttt{all\_levels}}
    \STATE \texttt{level.scaled\_score} $\leftarrow \left\lfloor S_{\max} \cdot \frac{\texttt{level.score} - \texttt{min\_global}}{\texttt{max\_global} - \texttt{min\_global}} \right\rfloor$
\ENDFOR
\STATE \textbf{return} \texttt{unified\_model}(\texttt{all\_levels}, \texttt{shap\_trees}, \texttt{top\_features})

\STATE \textbf{6. Prediction for New Sample $x^*$}
\STATE \texttt{total\_score} $\leftarrow 0$
\FOR{each feature $j$ in \texttt{top\_features}}
    \STATE \texttt{leaf} $\leftarrow T_j.\texttt{apply}(x^*_j)$
    \STATE \texttt{total\_score} $\leftarrow \texttt{total\_score} + \texttt{scaled\_score}[j][\texttt{leaf}]$
\ENDFOR
\STATE \textbf{return} \texttt{total\_score}
\end{algorithmic}
\end{algorithm*}
}

\section{Results}

To demonstrate the practical utility and advantages of the \texttt{SHAPoint} framework, we present an illustrative use-case focused on cardiovascular disease risk assessment (\Cref{subsec:use_case}). This example utilizes a publicly available dataset to showcase the end-to-end application of our proposed methodology, illustrating how \texttt{SHAPoint} can transform a high-performing but opaque "black-box" model into a simple, transparent, and clinically actionable risk scoring system, while preserving a significant portion of the original model's predictive power. This case study serves as an example of the framework's applicability to diverse datasets and is not intended to provide definitive clinical insights. Following the use-case, we present an in-depth comparison of SHAPoint with other cutting-edge methods in the field (\Cref{subsec:comparative}).

\subsection{Illustrative Use-case: Cardiovascular Disease Risk Assessment}
\label{subsec:use_case}
\subsubsection{Study Design and Data}
The study utilized the Cardiovascular Disease dataset from Kaggle \citep{sulianova2019cardiovascular}, which consists of 70,000 records of patient data with 11 features and a binary target variable indicating the presence or absence of cardiovascular disease. The features include: age (in days, converted to years), gender, height (cm), weight (kg), systolic blood pressure (ap\_hi), diastolic blood pressure (ap\_lo), cholesterol levels (1: normal, 2: above normal, 3: well above normal), glucose levels (1: normal, 2: above normal, 3: well above normal), smoking status, alcohol consumption, and physical a
ctivity. The dataset was split into training (80\%) and testing (20\%) sets, ensuring a stratified distribution of the target variable. We acknowledge that this dataset has been previously utilized in other research for similar illustrative purposes \citep{nguyen2019case, khan2020data}.

\subsubsection{Model Development}
First, a high-performance base ("black-box") model was developed using XGBoost. Hyperparameters were tuned using the Optuna framework to maximize AUC \citep{akiba2019optuna}. Next, we applied the \texttt{SHAPoint} framework. SHAP values were calculated from the base model to rank all features according to their relative importance. A parsimony analysis (shown in \Cref{fig:auc_complexity_cardio}) showed little to no improvement past 12 total parameters. Therefore, we chose to build the final risk score using the top \textbf{\textit{k=3}} features and a maximum of \textbf{4 leaves} per feature tree, as it offered a strong balance between relative performance and simplicity.

\subsubsection{Predictive Performance and Feature Importance}
\begin{figure}[t]
    \centering
    \includegraphics[width=0.9\textwidth]{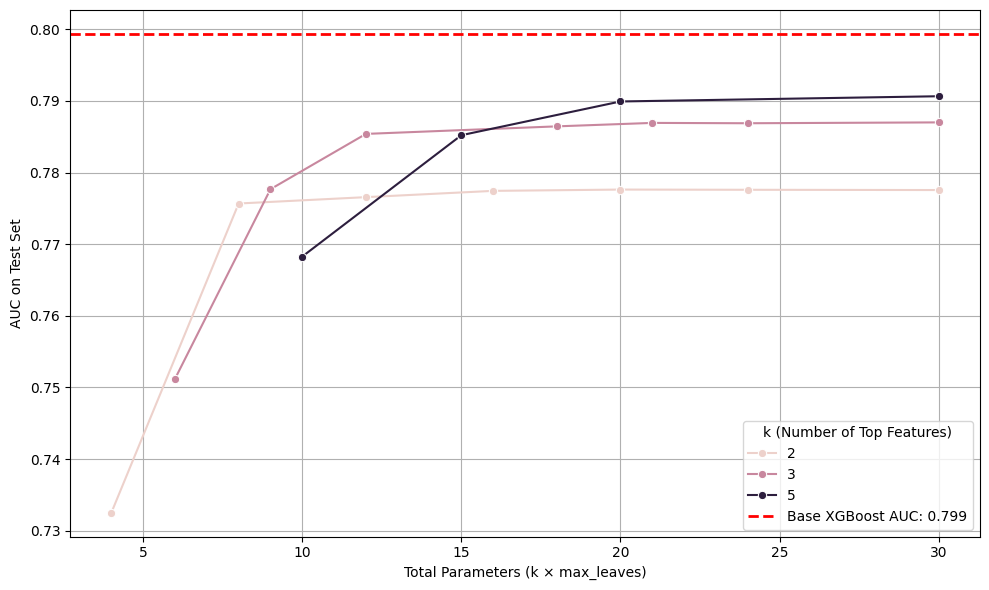}
    \caption{Area Under the Receiver Operating Characteristic Curve (AUC) versus Model Complexity on the test set for the Cardiovascular 
    Disease dataset. This plot illustrates the trade-off between predictive performance (AUC) and model complexity, defined as the total number of parameters (k top features × max leaves). The x-axis represents 'Total Parameters (k × max\_leaves)' and the y-axis represents 'AUC on Test Set'. Each colored line represents a different number of top features (k) used to build the \textit{SHAPoint} model.}
    \label{fig:auc_complexity_cardio}
\end{figure}

The optimized "black-box" XGBoost base model achieved a final \textbf{AUC of 0.799} on the test set. After applying the \texttt{SHAPoint} framework with the chosen configuration (\textit{k}=3, max leaves=4), the resulting interpretable risk score achieved a comparable \textbf{AUC of approximately 0.784} (\Cref{fig:auc_complexity_cardio}). This shows that our framework retained the vast majority of the base model's predictive power while drastically reducing its complexity.

\begin{figure}[t]
    \centering
 
    \includegraphics[width=0.7\textwidth]{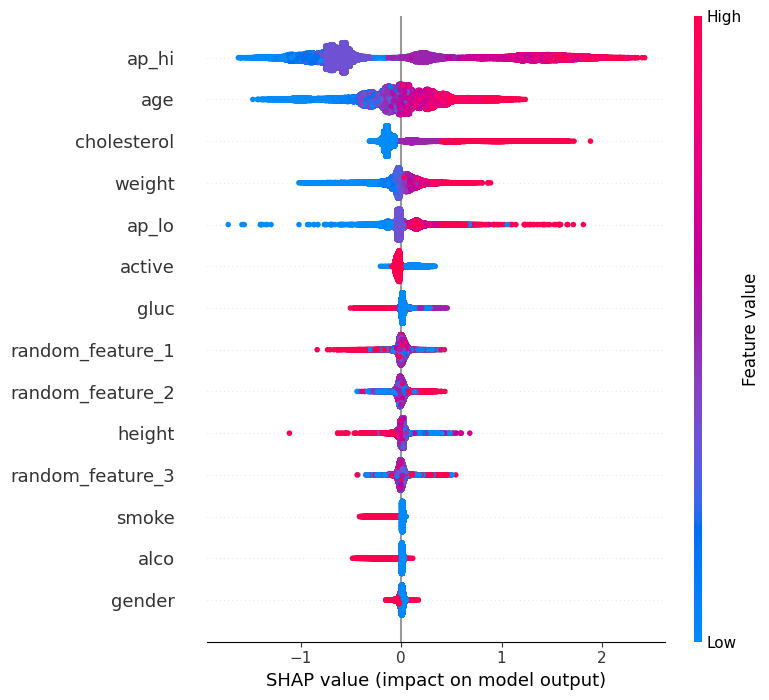}
    \caption{SHapley Additive exPlanations (SHAP) summary plot for the features of the Cardiovascular Disease prediction model, including injected randomized features. Each point on the beeswarm plot represents a single patient for a given feature. The position on the x-axis indicates the impact on the model's prediction (SHAP value), while the color represents the feature's value (red for high, blue for low). The x-axis represents 'SHAP value (impact on model output)' and the y-axis represents 'Feature Name'.}
    \label{fig:shap_beeswarm_cardio}
\end{figure}

T
he SHAP analysis identified the most impactful features for the model's predictions based on the provided dataset (\Cref{fig:shap_beeswarm_cardio}). The top 3 features selected for the final SHAPoint model were `ap\_hi` (systolic blood pressure), `age`, and `cholesterol`. This demonstrates that the framework effectively identifies and utilizes the most influential variables within the given data (\citep{kiran2025ai, han2024cardiovascular}. Notably, the beeswarm plot also includes random features, which serve to calibrate the feature selection process by ensuring that only features with predictive power exceeding that of random noise are chosen.

\subsubsection{Interpretable Risk Score and Validation}
\begin{figure}[t]
    \centering
    \includegraphics[width=\textwidth]{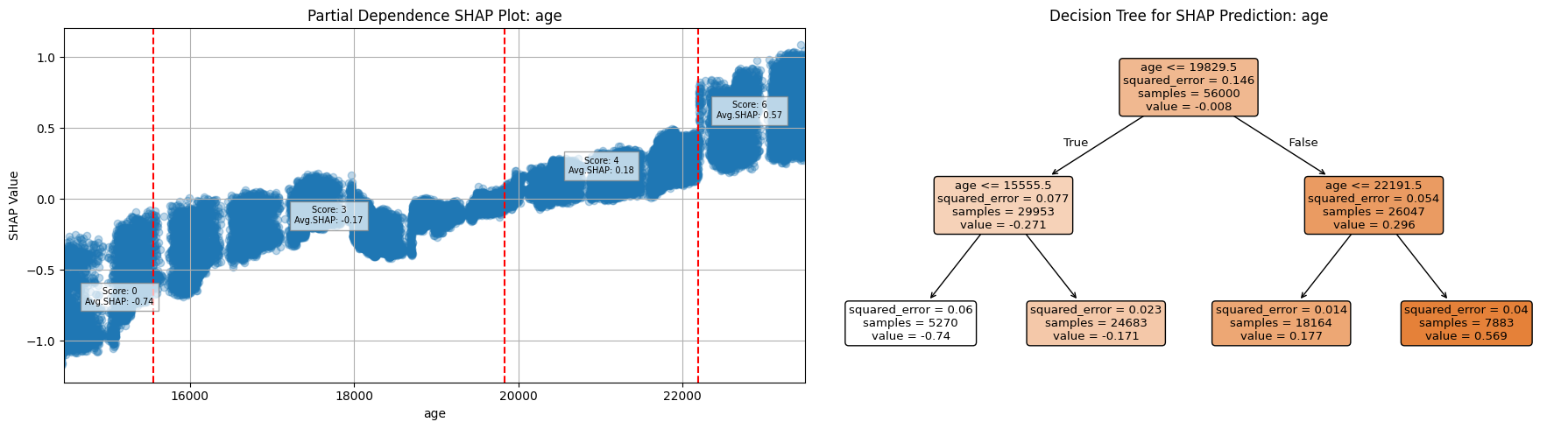}
    \caption{Visualization of the SHapley Additive exPlanations (SHAP)-based binning process for the age feature in the Cardiovascular Disease dataset. The left panel shows the partial dependence plot of SHAP values for age, illustrating how different ranges of the feature contribute to the model's output. The x-axis represents 'age' and the y-axis represents 'SHAP 
    Value (impact on model output)'. The right panel displays the decision tree trained to approximate these SHAP values, automatically identifying optimal split points and assigning integer scores to the resulting bins.}
    \label{fig:age_tree_cardio}
\end{figure}

The final output of the framework is a simple point-based risk score. The process for creating the score for o
ne of the selected features, `age`, is detailed in \Cref{fig:age_tree_cardio}. The decision tree automatically learns optimal splits to create bins, which are assigned integer scores. This data-driven process is repeated for all 3 features to generate a complete scoring table. The final output can be represented as a score sheet (seen in \Cref{tab:SHAPoint_scoring}), where each row is a rule for a single feature created by the framework.

\begin{figure}[htbp]
    \centering
    \includegraphics[width=0.9\textwidth]{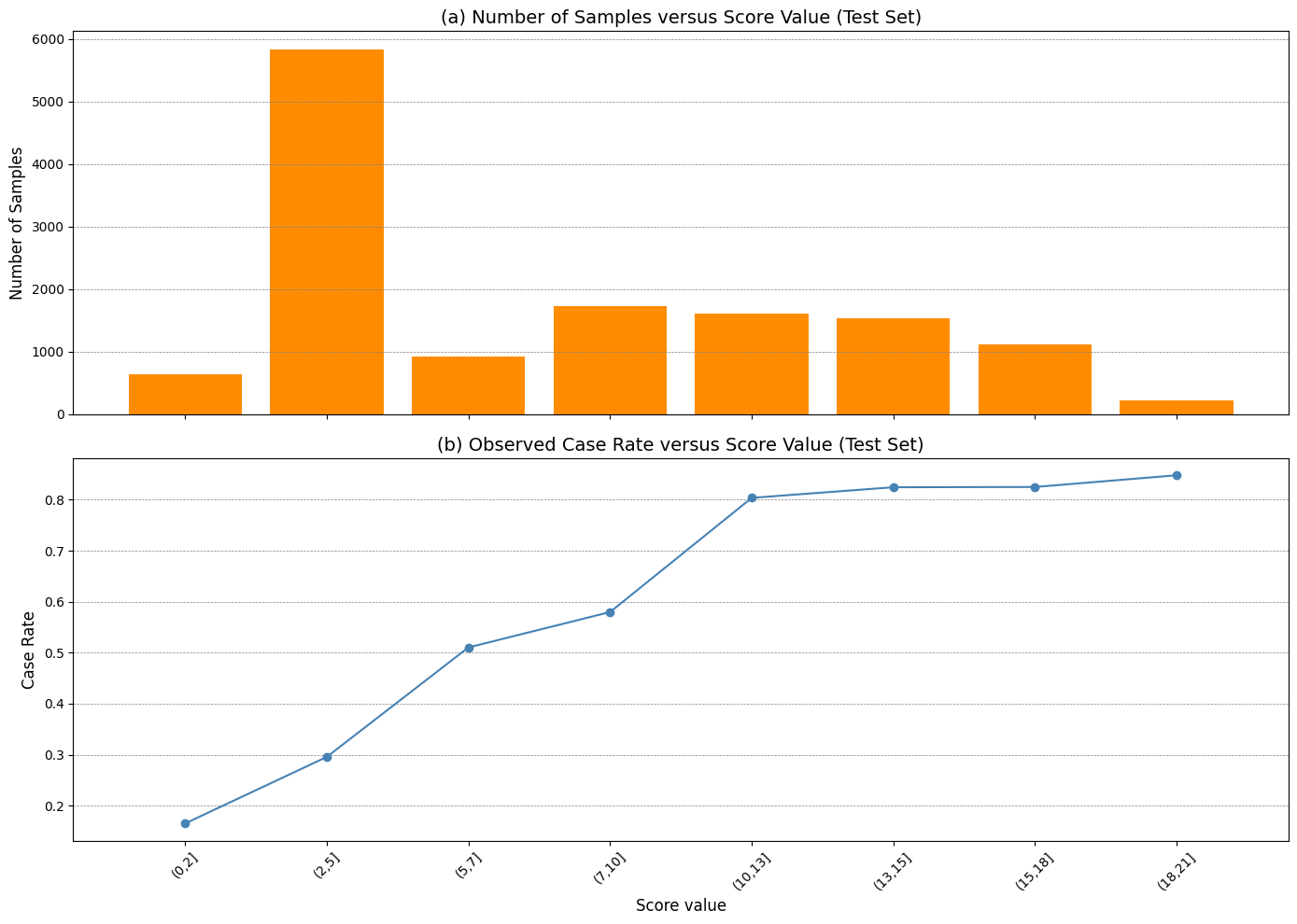}
    \caption{Clinical validation of the aggregated SHAPoint score for the Cardiovascular Disease dataset. Panel (a) shows the distribution of patients across different risk score bins on the test set, indicating that most patients fall into lower risk categories. The x-axis represents 'Score Value' bin and the y-axis represents 'Number of Samples'. Panel (b) illustrates the observed cardiovascular disease case rate as a function of the aggregated risk score, demonstrating a clear monotonic increase. The x-axis represents 'Score Value' bin and the y-axis represents 'Case Rate'.}
    \label{fig:score_distribution_cardio}
\end{figure}

The effectiveness of the final aggregated score in stratifying patients is demonstrated in \Cref{fig:score_distribution_cardio}. The distribution of patients across score bins shows that most individuals fall into low-risk categories (\Cref{fig:score_distribution_cardio}a). Crucially, the observed cardiovascular disease case rate increases monotonically with the risk score, rising from approximately 25\% to over 85\% (\Cref{fig:score_distribution_cardio}). This confirms the score's effectiveness in stratifying patients and its potential utility in a clinical setting for identifying high-risk individuals.

\begin{table}[htbp]
\caption{SHAPoint final scoring table for Cardi
ovascular disease prediction. Each row shows a rule for a single feature and its corresponding score by the model. The scoring table of the use case contains 11 rules, based on the three selected features. The final row shows the maximal accumulated score based on the rules.}
\centering
\begin{tabular}{@{}llr@{}}
\toprule
\textbf{Feature} & \textbf{Rule} & \textbf{Score} \\
\midrule
ap\_hi & ap\_hi $\leq$ 118.5 & 0 \\
 & ap\_hi $\leq$ 129.5  \&  ap\_hi $>$ 118.5 & 2 \\
 & ap\_hi $\leq$ 134.5  \&  ap\_hi $>$ 129.5 & 5 \\
 & ap\_hi $>$ 134.5 & 10 \\
age & age $\leq$ 42.5 & 0 \\
 & age $\leq$ 54.4  \&  age $>$ 42.5 & 3 \\
 & age $\leq$ 60.8  \&  age $>$ 54.4 & 4 \\
 & age $>$ 60.8 & 6 \\
cholesterol & cholesterol $\leq$ 1.5 & 0 \\
 & cholesterol $\leq$ 2.5  \&  cholesterol $>$ 1.5 & 1 \\
 & cholesterol $>$ 2.5 & 4 \\
\midrule
\multicolumn{2}{l}{\textbf{Total Potential Score}} & \textbf{20} \\
\bottomrule

\end{tabular}

\small

\label{tab:SHAPoint_scoring}
\end{table}

\subsubsection{Interactive Tool}
\begin{figure}[t]
    \centering
    \includegraphics[width=\textwidth]{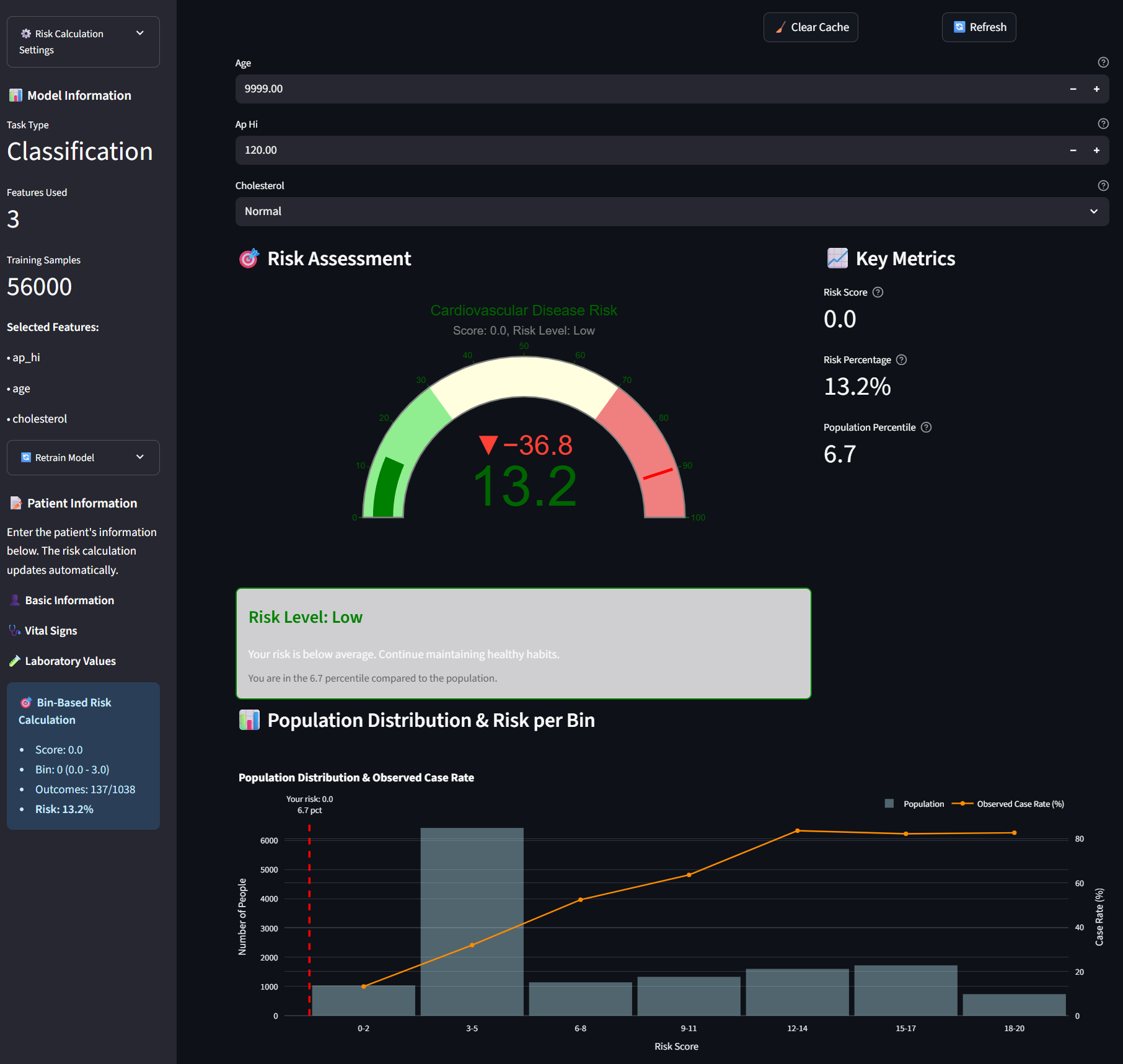} 
    \caption{A preview of the SHAPoint web application, illustrating an interactive interface for risk assessment. This figure shows the user interface of the application, including input fields for patient features (e.g., Age, Ap-Hi, Cholesterol) and the resulting risk assessment (Risk Score, Risk Percentage, Population Percentile, Risk Level). The bottom panel displays a distribution of the population and observed case rate by risk bin, with the x-axis representing 'Risk Score' and the y-axis representing 'Number of People' (left) and 'Observed Case Rate (\%)' (right).}
    \label{fig:webapp_page}
\end{figure}

Beyond the core methodology, we have also initiated the development of a Streamlit-based web application to provide an interactive demonstration of the capabilities of the SHAPoint framework, presented in \Cref{fig:webapp_page} \cite{streamlit2024}. The tool currently supports classification tasks and can be further configured (e.g., training additional models, feature filtering, adjusting scales, and presentation configuration). This tool aims to offer a user-friendly interface for exploring interpretable risk assessment, showcasing the potential for broader accessibility and practical deployment of such models in the future. 

\subsubsection{Comparison with FasterRisk}
To illustrate the utility of SHAPoint in a practical scenario, we conducted a targeted comparison with FasterRisk using the Cardiovascular Disease dataset. While a detailed methodological comparison and extensive benchmarking against FasterRisk and other state-of-the-art methods are presented in the Comparative Study section, this analysis further emphasizes SHAPoint's competitive predictive performance, interpretability, and computational efficiency. Specifically, SHAPoint achieved an ROC AUC of 0.784 and a PR AUC of 0.751, requiring only 16.75 seconds for training with hyperparameter optimization, and 9.52 seconds without optimization. By contrast, FasterRisk's original implementation (without binarization) obtained an ROC AUC of 0.594 in 56.46 seconds. When FasterRisk utilized its built-in binarization utility—configured to create thresholds equal to the total number of parameters—it achieved comparable predictive performance (ROC AUC 0.774, PR AUC 0.744) but required significantly longer training time (173.72 seconds). These findings underscore SHAPoint's advantages in terms of flexibility, interpretability, and efficiency.

\subsection{Comparative Study}
\label{subsec:comparative}

To rigorously benchmark SHAPoint, we selected state-of-the-art baselines (FasterRisk, ShapleyVIC, and AutoScore-Survival) widely recognized for their interpretability, predictive accuracy, and computational efficiency. For classification tasks, we compare SHAPoint with FasterRisk and ShapleyVIC, which have previously demonstrated superior accuracy and runtime efficiency. For survival analysis tasks, we benchmark against the AutoScore-Survival method. All evaluations are conducted on publicly available healthcare datasets, as simple scoring systems are commonly employed in clinical settings for ease of interpretation and implementation. Furthermore, a supplementary analysis against a baseline logistic regression model is provided in the Supplementary Materials to contextualize performance against traditional explainable methods.

For our comparative analysis, we utilized six publicly available datasets. The classification tasks were evaluated on the Cardiovascular Disease dataset from Kaggle \citep{sulianova2019cardiovascular}, the Breast Cancer dataset from the scikit-learn library \citep{scikit-learn}, and the MIMIC-III ICU clinical database \citep{johnson2016mimicphysionet, johnson2016mimicscientific}. For the survival analysis tasks, we employed a curated Pan-Cancer dataset from The Cancer Genome Atlas (TCGA) \citep{liu2018integrated}, along with the Lung Cancer and Breast Cancer datasets provided by the scikit-survival library \citep{scikit-survival}. 

We used the Area Under the ROC Curve (AUC) as the evaluation metric for classification tasks, and the concordance index (C-index) for survival analysis. Predictive performance was estimated using a stratified 10-fold cross-validation, repeated five times to ensure robustness. For fair comparison, we first executed each baseline method using their default hyperparameters. Then, we configured the SHAPoint model to achieve comparable model complexity, defined by the number of lines, which refers to the number of features and the number of levels (cutoff points or bins) in the resulting scorecard. The internal XGBoost model within SHAPoint was also trained using XGBoost’s default hyperparameters. While further gains can be achieved through hyperparameter optimization (e.g., via Bayesian optimization), the current implementation allows users to employ Optuna for fine-tuning the XGBoost component if desired. 

We used the Area Under the ROC Curve (AUC) as the evaluation metric for classification tasks, and the concordance index (C-index) for survival analysis. Predictive performance was estimated using a stratified 10-fold cross-validation, repeated five times to ensure robustness. For fair comparison, we first executed each baseline method using their default hyperparameters. Then, we configured the SHAPoint model to achieve comparable model complexity, defined by the number of lines, which refers to the number of features and the number of levels (cutoff points or bins) in the resulting scorecard. In addition, we reported the average running time. All experiments were executed on a single thread of an Intel i7-based machine to provide consistent runtime comparisons. 

For the comparison with FasterRisk, we evaluated two versions of SHAPoint: a default version with standard parameters and an Optuna-optimized version. Similarly, for FasterRisk, we included two versions: one fitting the optimizer without pre-binning and another using its internal binarization tool (binarizer) with the number of bins set to match the number of leaves. This comprehensive comparison allows for a more robust analysis of their performance across different complexities and configurations, including PR AUC as an additional evaluation metric. For the comparisons with ShapleyVIC and Autoscore-Survival), the internal XGBoost model within SHAPoint was trained using XGBoost’s default hyperparameters. 

\subsubsection{Comparison Results}

\Cref{tab:cardiovascular_comparison}, \Cref{tab:breastcancer_comparison}, and \Cref{tab:mimic_comparison} summarize the detailed comparison between SHAPoint and FasterRisk across various complexities for the Cardiovascular Disease, Breast Cancer, and MIMIC-III ICU datasets, respectively. \Cref{tab:classification_shapleyvic} and \Cref{tab:survival} present comparisons with ShapleyVIC and AutoScore-Survival. This separation is necessary because each baseline method results in different model complexities.

Across all datasets and complexities, SHAPoint consistently demonstrates superior or comparable predictive performance (both ROC AUC and PR AUC) while maintaining significantly faster training times compared to FasterRisk. Notably, even the default version of SHAPoint often outperforms FasterRisk, and the Optuna-optimized SHAPoint consistently achieves the best performance among all compared models in most scenarios. For instance, on the Cardiovascular Disease dataset, SHAPoint (Default) achieves an ROC AUC of $0.770 \pm 0.004$ and a PR AUC of $0.733 \pm 0.005$ with a training time of $3.81 \pm 0.08$ seconds for a complexity of 6, while FasterRisk yields an ROC AUC of $0.696 \pm 0.041$ (or 0.582 without binarization) and a PR AUC of $0.659 \pm 0.036$ ( with a training time of $11.15 \pm 0.74$ seconds. When FasterRisk utilizes its internal binarizer, its training time can be substantially higher, as seen in the Cardiovascular Disease dataset at complexity 20 ($264.83 \pm 67.41$ seconds for FasterRisk with binarizer vs. $5.10 \pm 0.10$ seconds for SHAPoint Default).

On the Breast Cancer dataset, both SHAPoint versions generally outperform FasterRisk in terms of training time, while achieving comparable or better predictive performance. For example, at complexity 6, SHAPoint (Default) has an ROC AUC of $0.970 \pm 0.013$ and a PR AUC of $0.973 \pm 0.013$ with a training time of $0.21 \pm 0.01$ seconds, whereas FasterRisk (without binarization) has an ROC AUC of $0.973 \pm 0.016$ but a training time of $2.10 \pm 0.07$ seconds. At higher complexities, FasterRisk with binarizer can achieve slightly higher AUC/PR AUC, but at a significantly increased training time.

For the MIMIC-III ICU dataset, SHAPoint (Default) and SHAPoint (Optuna) generally show competitive performance with FasterRisk, particularly at higher complexities, while maintaining a substantial advantage in training time. For instance, at complexity 20, SHAPoint (Optuna) achieves an ROC AUC of $0.838 \pm 0.015$ and a PR AUC of $0.332 \pm 0.039$ with a training time of $7.25 \pm 0.75$ seconds, whereas FasterRisk (with binarizer) yields an ROC AUC of $0.820 \pm 0.014$ and a PR AUC of $0.299 \pm 0.028$ with a training time of $333.85 \pm 17.01$ seconds. The user's observation that FasterRisk slightly outperforms SHAPoint on MIMIC with 2 features (complexity 6) is noted, and can be attributed to FasterRisk's ability to select up to 6 different features (k=6) in that complexity setting, while SHAPoint was limited to two features and three bins, highlighting the impact of feature selection flexibility.

In comparison with ShapleyVIC and AutoScore-Survival, SHAPoint continues to show competitive predictive performance and superior computational efficiency, as detailed in \Cref{tab:classification_shapleyvic} and \Cref{tab:survival}. SHAPoint consistently achieves faster runtimes using a single core. Given that SHAPoint leverages XGBoost, its execution time can be further improved through parallel processing (multi-threading, GPU acceleration, or distributed computing) as well as by utilizing optimized algorithms such as histogram-based or approximate greedy splitting. However, to ensure fair comparison we have not used all these acceleration options.

It is important to note that the final complexity of models generated by SHAPoint may be lower than the user-specified limits. This is because the user can constrain the number of variables and the number of levels per variable independently, but SHAPoint applies internal pruning that may reduce the number of levels for each variable. As a result, the final model may be more concise than initially intended, improving interpretability without sacrificing performance.

\begin{table*}[htbp]
\caption{Comparison of \textbf{SHAPoint} and \textbf{FasterRisk} for the Cardiovascular Disease dataset (original shape: 70000 $\times$ 11).}
\centering
\renewcommand{\arraystretch}{1.2}
\resizebox{\textwidth}{!}{%
\begin{tabular}{p{3.8cm}llllll}
\toprule
Complexity & Model & ROC AUC & PR AUC & Training Time \\
\midrule
6 (2 features, 3 leaves or k=6) & SHAPoint (Default) & $0.770 \pm 0.004$ & $0.733 \pm 0.005$ & \bm{$3.81 \pm 0.08$} \\
~ & SHAPoint (Optuna) & \bm{$0.771 \pm 0.004$} & \bm{$0.735 \pm 0.004$} & $13.01 \pm 0.98$ \\
~ & FasterRisk & $0.582 \pm 0.041$ & $0.575 \pm 0.038$ & $43.31 \pm 0.57$ \\
~ & FasterRisk (Binarizer=3) & $0.696 \pm 0.041$ & $0.659 \pm 0.036$ & $11.15 \pm 0.74$ \\
\cmidrule(l){1-5}
12 (3 features, 4 leaves or k=12) & SHAPoint (Default) & \bm{$0.789 \pm 0.003$} & $0.76 \pm 0.004$ & \bm{$4.57 \pm 0.07$} \\
~ & SHAPoint (Optuna) & \bm{$0.789 \pm 0.003$} & \textbf{$0.761 \pm 0.004$} & $12.06 \pm 1.27$ \\
~ & FasterRisk & $0.596 \pm 0.004$ & $0.579 \pm 0.006$ & $50.05 \pm 1.43$ \\
~ & FasterRisk (Binarizer=4) & $0.745 \pm 0.037$ & $0.716 \pm 0.032$ & $65.81 \pm 7.5$ \\
\cmidrule(l){1-5}
20 (5 features, 4 leaves or k=20) & SHAPoint (Default) & $0.792 \pm 0.002$ & \textbf{$0.768 \pm 0.004$} & \bm{$5.1 \pm 0.1$} \\
~ & SHAPoint (Optuna) & \bm{$0.793 \pm 0.002$} & \bm{$0.768 \pm 0.003$} & $12.91 \pm 1.8$ \\
~ & FasterRisk & $0.596 \pm 0.004$ & $0.579 \pm 0.006$ & $49.98 \pm 1.41$ \\
~ & FasterRisk (Binarizer=4) & $0.778 \pm 0.009$ & $0.743 \pm 0.016$ & $264.83 \pm 67.41$ \\
\bottomrule
\end{tabular}%
}

\label{tab:cardiovascular_comparison}
\end{table*}

\begin{table*}[htbp]
\caption{Comparison of \textbf{SHAPoint} and \textbf{FasterRisk} for the Breast Cancer dataset (original shape: 569 $\times$ 30).}
\centering
\renewcommand{\arraystretch}{1.2}
\resizebox{\textwidth}{!}{%
\begin{tabular}{p{3.8cm}llllll}
\toprule
Complexity & Model & ROC AUC & PR AUC & Training Time \\
\midrule
6 (2 features, 3 leaves or k=6) & SHAPoint (Default) & $0.970 \pm 0.013$ & $0.973 \pm 0.013$ & \bm{$0.21 \pm 0.01$} \\
~ & SHAPoint (Optuna) & $0.960 \pm 0.023$ & $0.962 \pm 0.025$ & $3.96 \pm 0.34$ \\
~ & FasterRisk & \bm{$0.973 \pm 0.016$} & \bm{$0.985 \pm 0.009$} & $2.1 \pm 0.07$ \\
~ & FasterRisk (Binarizer=3) & \bm{$0.973 \pm 0.014$} & $0.978 \pm 0.008$ & $1.63 \pm 0.1$ \\
\cmidrule(l){1-5}
12 (3 features, 4 leaves or k=12) & SHAPoint (Default) & $0.978 \pm 0.014$ & $0.98 \pm 0.016$ & \bm{$0.41 \pm 0.01$} \\
~ & SHAPoint (Optuna) & $0.969 \pm 0.011$ & $0.971 \pm 0.014$ & $4.16 \pm 0.19$ \\
~ & FasterRisk & $0.953 \pm 0.025$ & $0.974 \pm 0.013$ & $8.21 \pm 0.26$ \\
~ & FasterRisk (Binarizer=4) & \bm{$0.989 \pm 0.005$} & \bm{$0.992 \pm 0.003$} & $6.67 \pm 0.39$ \\
\cmidrule(l){1-5}
20 (5 features, 4 leaves or k=20) & SHAPoint (Default) & $0.984 \pm 0.012$ & $0.984 \pm 0.014$ & \bm{$0.61 \pm 0.02$} \\
~ & SHAPoint (Optuna) & $0.986 \pm 0.011$ & $0.988 \pm 0.01$ & $4.32 \pm 0.11$ \\
~ & FasterRisk & $0.928 \pm 0.064$ & $0.954 \pm 0.045$ & $19.93 \pm 0.39$ \\
~ & FasterRisk (Binarizer=4) & \bm{$0.992 \pm 0.007$} & \bm{$0.995 \pm 0.005$} & $18.37 \pm 0.5$ \\
\bottomrule
\end{tabular}%
}

\label{tab:breastcancer_comparison}
\end{table*}

\begin{table*}[htbp]
\caption{Comparison of \textbf{SHAPoint} and \textbf{FasterRisk} for the MIMIC-III ICU dataset (original shape: 14,000 $\times$ 20).}
\centering
\renewcommand{\arraystretch}{1.2}
\resizebox{\textwidth}{!}{%
\begin{tabular}{p{3.8cm}llllll}
\toprule
Complexity & Model & ROC AUC & PR AUC & Training Time \\
\midrule
6 (2 features, 3 leaves or k=6) & SHAPoint (Default) & $0.714 \pm 0.016$ & $0.155 \pm 0.01$ & \bm{$0.94 \pm 0.02$} \\
~ & SHAPoint (Optuna) & $0.719 \pm 0.008$ & $0.167 \pm 0.006$ & $7.41 \pm 1.59$ \\
~ & FasterRisk & \bm{$0.773 \pm 0.026$} & \bm{$0.243 \pm 0.033$} & $18.75 \pm 0.81$ \\
~ & FasterRisk (Binarizer=3) & $0.717 \pm 0.015$ & $0.169 \pm 0.013$ & $18.65 \pm 1.29$ \\
\cmidrule(l){1-5}
12 (3 features, 4 leaves or k=12) & SHAPoint (Default) & $0.780- \pm 0.018$ & $0.233 \pm 0.015$ & \bm{$1.22 \pm 0.02$} \\
~ & SHAPoint (Optuna) & $0.782 \pm 0.013$ & $0.234 \pm 0.016$ & $6.98 \pm 0.75$ \\
~ & FasterRisk & \bm{$0.799 \pm 0.021$} & \bm{$0.295 \pm 0.033$} & $75.31 \pm 1.23$ \\
~ & FasterRisk (Binarizer=4) & $0.798 \pm 0.018$ & $0.258 \pm 0.02$ & $98.03 \pm 6.75$ \\
\cmidrule(l){1-5}
20 (5 features, 4 leaves or k=20) & SHAPoint (Default) & $0.837 \pm 0.019$ & $0.331 \pm 0.03$ & \bm{$1.53 \pm 0.04$} \\
~ & SHAPoint (Optuna) & \bm{$0.838 \pm 0.015$} & \bm{$0.332 \pm 0.039$} & $7.25 \pm 0.75$ \\
~ & FasterRisk & $0.785 \pm 0.019$ & $0.275 \pm 0.021$ & $125.27 \pm 2.38$ \\
~ & FasterRisk (Binarizer=4) & $0.820 \pm 0.014$ & $0.299 \pm 0.028$ & $333.85 \pm 17.01$ \\
\bottomrule
\end{tabular}%
}
\label{tab:mimic_comparison}
\end{table*}

\begin{table*}[ht]
\caption{Classification Task: Comparing SHAPoint to ShapleyVIC (via AutoScore).}
\centering
\newcolumntype{C}[1]{>{\centering\arraybackslash}p{#1}}
\resizebox{\textwidth}{!}{%
  \begin{tabular}{p{3.5cm} c C{1.6cm} C{1.6cm} c C{1.6cm} C{1.6cm}}
    \toprule
    Dataset & \multicolumn{3}{c}{ShapleyVIC with AutoScore} & \multicolumn{3}{c}{SHAPoint} \\
    & AUC & Running \newline Time & Complexity \newline (lines) & AUC & Running \newline Time & Complexity \newline (lines) \\
    \midrule
    Cardiovascular Disease (Kaggle) & $0.784 \pm 0.07$ & 12.11 & 40 & \bm{$0.787 \pm 0.05$} & \bm{$9.63$}  & 36 \\
    Breast Cancer (scikit-learn)    & \bm{$0.99 \pm 0.01$}  & 2.61  & 16 & \bm{$0.99 \pm 0.06$}  & \bm{$1.05$}  & 16 \\
    MIMIC-III ICU                   & $0.826 \pm 0.12$ & 24.95 & 30 & \bm{$0.829 \pm 0.09$} & \bm{$18.77$} & 26 \\
    \bottomrule
  \end{tabular}%
}

\label{tab:classification_shapleyvic}
\end{table*}

\begin{table*}[ht]
\caption{Survival Analysis Task: Comparing SHAPoint to AutoScore-Survival. Dataset details (samples, features): PanCan (TCGA) (1,053, 1,000); Lung Cancer (sksurv) (137, 6); Breast Cancer (sksurv) (198, 80).}
\centering
\newcolumntype{C}[1]{>{\centering\arraybackslash}p{#1}}
\resizebox{\textwidth}{!}{%
  \begin{tabular}{p{3.5cm} c C{1.6cm} C{1.6cm} c C{1.6cm} C{1.6cm}}
    \toprule
    Dataset & \multicolumn{3}{c}{AutoScore-Survival} & \multicolumn{3}{c}{SHAPoint} \\
    & C-Index & Running \newline Time & Complexity \newline (lines) & C-Index & Running \newline Time & Complexity \newline (lines) \\
    \midrule
    PanCan (TCGA)              & $0.679 \pm 0.09$ & 22.51 & 8 & \bm{$0.696 \pm 0.12$} & \bm{$13.78$} & 8 \\
    Lung Cancer (sksurv)       & \bm{$0.721 \pm 0.06$} & 15.73 & 6 & $0.717 \pm 0.07$ & \bm{$9.01$}  & 6 \\
    Breast Cancer (sksurv)     & $0.873 \pm 0.09$ & 13.40 & 8 & \bm{$0.875 \pm 0.07$} & \bm{$6.80$}  & 8 \\
    \bottomrule
  \end{tabular}%
}

\label{tab:survival}
\end{table*}

\Cref{tab:survival} summarizes the results of the survival analysis task comparing SHAPoint and AutoScore-Survival. While both methods achieved comparable C-index performance, SHAPoint consistently demonstrated lower computational time, highlighting its efficiency advantage.

\section{Discussion}

The proposed model provides several advantages for risk assessment applications. SHAPoint model works equally well for classification, regression, and survival analysis, as SHAP and XGBoost are not restricted to a specific loss function. This overcomes the task-specific design of AutoScore (classification) and AutoScore-Survival (survival only). Moreover, unlike AutoScore, SHAPoint uses decision trees to partition variables based on SHAP prediction, avoiding the need for human-defined or quantile-based cutoffs. This makes variable binning data-driven and adaptive, improving both interpretability and fidelity. Because binning is directly based on the trained XGBoost model’s SHAP values, this ensures the score representation reflects the base model’s learned structure, unlike AutoScore which uses logistic regression scores after a random forest selection step.

Moreover, because XGBoost and TreeSHAP are highly optimized for speed and scalability, SHAPoint avoids the computational bottleneck of ShapleyVIC, which samples and analyzes hundreds of models. Finally, by relying on XGBoost, SHAPoint inherits native support for handling missing values and enforcing monotonic constraints—features not inherently present in logistic regression–based pipelines like AutoScore. \Cref{tab:comparison} summarizes the differences between the proposed method and existing state-of-the-art risk modeling methods. As demonstrated in our supplementary analysis (Supplementary Table 1), this approach leads to superior performance, especially in PR-AUC, compared to a standard logistic regression model, while maintaining a competitive ROC-AUC. Furthermore, the per-sample nature of SHAP values provides unique flexibility for post-hoc analysis; specialized risk scores can be generated for any subpopulation (e.g., younger individuals) by isolating their corresponding SHAP values and reapplying the scoring procedure, all without having to retrain the base model.

The comprehensive comparison with FasterRisk, including both default and optimized versions of SHAPoint and FasterRisk (with and without binarization), further underscores SHAPoint's advantages. Even without explicit optimization, SHAPoint consistently delivers strong predictive performance (both ROC AUC and PR AUC) with significantly shorter training times across various complexities and datasets. For larger complexities and datasets, SHAPoint often outperforms FasterRisk, and when it doesn't, it remains on-par in terms of predictive performance while maintaining a substantial advantage in training speed. The specific case of MIMIC with complexity 6 (2 features), where FasterRisk shows better performance, can be attributed to FasterRisk's ability to select up to 6 different features (k=6) in that complexity setting, whereas SHAPoint was limited to two features and three bins, highlighting the impact of feature selection flexibility. 

While SHAPoint offers substantial improvements in interpretability and clinical utility, it is important to acknowledge certain inherent limitations related to its reliance on SHAP approximations. The fidelity of the simplified risk scores to the underlying XGBoost model depends on SHAP's additive feature attribution, which may not always fully capture complex nonlinearities or nuanced higher-order interactions \citep{kumar2020problems}. Recent studies suggest that SHAP explanations can be less precise in scenarios involving particularly strong feature interactions, potentially offering incomplete representations of model behavior \citep{sundararajan2020many, ning2022shapley}. Therefore, in clinical contexts marked by intricate feature synergies, SHAPoint balances some degree of fidelity for the benefit of enhanced interpretability and usability.

The case study on the Cardiovascular Disease dataset demonstrates that the proposed methodology efficiently generates interpretable risk scores, suitable for practical application within the context of the provided data. The selected features (`ap\_hi`, `age`, `cholesterol`) are consistent with common risk factors for cardiovascular disease, illustrating the model's interpretability and its ability to identify influential variables within the dataset. The transparent interpretability provided by decision-tree binning facilitates understanding of the model's decision process by clearly defining risk thresholds.

This analysis highlights SHAPoint's capability to derive risk scores from complex datasets, demonstrating performance and computational efficiency competitive with established methods such as FasterRisk. Its unique SHAP-driven approach provides a robust and transparent bridge from complex machine learning models to simple, effective scoring tools. 

It is important to note that while the results are promising for illustrative purposes, this study does not constitute a clinical recommendation, and further validation with clinically curated datasets would be necessary for real-world medical application.

Key insights derived include the significant predictive power of easily accessible parameters within the dataset, highlighting the potential for routine screening and monitoring. Further, the efficient complexity-performance balance positions this method favorably relative to existing risk-scoring approaches. 

Future work could focus on exploring the application of SHAPoint in real-world clinical settings with large Electronic Health Record (EHR) systems. This would involve addressing challenges related to data integration, scalability, and the dynamic nature of clinical data. Furthermore, enhancing the flexibility of rounding the cutoffs for continuous values could further simplify the use of SHAPoint for clinicians, facilitating its seamless integration into clinical workflows \citep{molero2024optimal}.  

\begin{table*}[htpb]
\caption{Summary of Main Features of State-of-the-Art Algorithms}
\centering
\resizebox{\textwidth}{!}{%
  \begin{tabular}{lccccc}
    \toprule
    Criterion                             & SHAPoint & AutoScore & FasterRisk & ShapleyVIC & RiskSLIM \\
    \midrule
    Supports survival tasks               & \cmark    & \cmark\,(via AutoScore-Survival) & \xmark   & \xmark   & \xmark   \\
    Supports regression tasks             & \cmark    & \xmark    & \xmark   & \xmark   & \xmark   \\
    No need to define cutoffs for numeric vars
                                          & \cmark    & \xmark    & \xmark   & \cmark   & \cmark   \\
    Model-aligned variable binning        & \cmark    & \xmark    & \xmark   & \xmark   & \xmark   \\
    Running time                          & Ultra-Low & Moderate  & Low      & Moderate & High     \\
    Handles missing values natively       & \cmark    & \xmark    & \xmark   & \xmark   & \xmark   \\
    Supports monotonic constraints        & \cmark    & \xmark    & \xmark   & \xmark   & \cmark   \\
    Flexible and extendable               & \cmark    & Moderate  & Low      & Moderate & \xmark   \\
    \bottomrule
  \end{tabular}%
}

\label{tab:comparison}
\end{table*}

\bibliographystyle{acm}
\bibliography{ref}

\end{document}


\section*{Supplementary Materials}

This section provides supplementary details regarding the publicly available datasets utilized in this study.

\subsection*{Cardiovascular Disease Dataset}

The Cardiovascular Disease Dataset, available on Kaggle \cite{sulianova2019cardiovascular}, is a widely used dataset for classification tasks aimed at predicting the presence of cardiovascular disease. This dataset typically includes features such as age, gender, cholesterol levels, blood pressure, and other relevant medical indicators. It is a valuable resource for developing and testing machine learning models for heart disease prediction.

\subsection*{MIMIC-III Clinical Database Derived Datasets}

The Medical Information Mart for Intensive Care (MIMIC-III) Clinical Database \cite{johnson2016mimicphysionet, johnson2016mimicscientific} is a large, freely available database containing de-identified health-related data from over forty thousand patients admitted to critical care units at the Beth Israel Deaconess Medical Center between 2001 and 2012. It integrates a wide array of data, including demographics, vital signs, laboratory test results, procedures, medications, caregiver notes, and imaging reports. For this study, we derived \textbf{Mortality Prediction Dataset}. This dataset supports a survival analysis task, aiming to predict in-hospital mortality. It includes patient characteristics, severity scores, and time-to-event information relevant to mortality outcomes.

Access to the MIMIC-III database requires completion of a data use agreement via \href{https://physionet.org/content/mimiciii/1.4/}{PhysioNet}. All data within MIMIC-III are de-identified in accordance with Health Insurance Portability and Accountability Act (HIPAA) standards, with dates shifted to preserve intervals while protecting patient privacy. Further details on the de-identification process and database structure can be found in the original MIMIC-III publication \cite{johnson2016mimicscientific} and on the PhysioNet website.

\subsection*{TCGA Pan-Cancer Dataset}

The Cancer Genome Atlas (TCGA) Pan-Cancer dataset \cite{liu2018integrated} provides a comprehensive resource for cancer research, integrating clinical data with genomic and molecular information across various cancer types. This dataset is instrumental for survival outcome analytics and understanding the molecular underpinnings of cancer. It includes patient demographics, tumor characteristics, treatment details, and follow-up information, making it suitable for a wide range of prognostic and predictive modeling tasks in oncology.

\subsection*{Scikit-learn and Scikit-survival Datasets}

For demonstrating the versatility of SHAPoint across different prediction tasks, we also utilized several built-in datasets from popular Python libraries. The \textbf{breast cancer dataset} from scikit-learn \cite{scikit-learn} was used for classification tasks. Additionally, datasets from scikit-survival \cite{scikit-survival}, including specific \textbf{lung cancer} and \textbf{breast cancer datasets}, were employed for survival analysis tasks. These datasets provide readily accessible and well-structured data for benchmarking and illustrating the application of SHAPoint in diverse medical contexts.

\subsection*{Comparison with Logistic Regression}
To further contextualize the performance of SHAPoint, we conducted a supplementary analysis comparing it to a standard logistic regression model, which is a widely recognized explainable baseline. While logistic regression offers some interpretability, it lacks the advanced features of SHAPoint, such as automated data-driven variable binning and native handling of non-linear relationships. The results, presented in Supplementary Table~\ref{suptab:model_comparison}, demonstrate that SHAPoint achieves comparable or superior predictive performance, particularly in PR-AUC, across different feature counts on both the Cardiovascular and MIMIC datasets. This comparison highlights the ability of SHAPoint to maintain high predictive accuracy while offering enhanced interpretability and flexibility over traditional linear models.

\begin{table}[htbp]
\centering
\caption{Performance comparison between SHAPoint and Logistic Regression models across different feature counts}
\label{suptab:model_comparison}
\begin{tabular}{cccccc}
\toprule
\multirow{6}{*}{\textbf{Dataset}} & \multirow{3}{*}{\textbf{Model}} & \textbf{Features} & \textbf{ROC-AUC} & \textbf{PR-AUC} \\
\midrule
\multirow{6}{*}{Cardiovascular} & \multirow{3}{*}{SHAPoint} & 3 & 0.788 $\pm$ 0.003 & 0.757 $\pm$ 0.004 \\
 &  & 5 & 0.791 $\pm$ 0.003 & 0.761 $\pm$ 0.004 \\
 &  & 10 & 0.793 $\pm$ 0.003 & 0.763 $\pm$ 0.004 \\
 & \multirow{3}{*}{LogisticRegression} & 3 & 0.782 $\pm$ 0.004 & 0.662 $\pm$ 0.004 \\
 &  & 5 & 0.783 $\pm$ 0.003 & 0.662 $\pm$ 0.002 \\
 &  & 10 & 0.784 $\pm$ 0.004 & 0.664 $\pm$ 0.002 \\
\midrule
\multirow{6}{*}{MIMIC} & \multirow{3}{*}{SHAPoint} & 3 & 0.782 $\pm$ 0.007 & 0.228 $\pm$ 0.011 \\
 &  & 5 & 0.823 $\pm$ 0.007 & 0.307 $\pm$ 0.020 \\
 &  & 10 & 0.845 $\pm$ 0.006 & 0.353 $\pm$ 0.020 \\
 & \multirow{3}{*}{LogisticRegression} & 3 & 0.801 $\pm$ 0.008 & 0.104 $\pm$ 0.010 \\
 &  & 5 & 0.854 $\pm$ 0.009 & 0.154 $\pm$ 0.011 \\
 &  & 10 & 0.868 $\pm$ 0.007 & 0.181 $\pm$ 0.009 \\
\bottomrule
\end{tabular}

\end{table}

\bibliographystyle{naturemag}
\bibliography{ref}


\section*{Code Availability}

\noindent
To allow reproducibility and ease of use, the code for SHAPoint model and SHAPoint webapp are both publicly available on GitHub repositories:
\begin{enumerate}
    \item \href{https://github.com/TomerMeirman/SHAPoint}{SHAPoint - Risk Model GitHub Repository}
    \item \href{https://github.com/TomerMeirman/SHAPoint_webapp}{SHAPoint - Webapp GitHub Repository}
\end{enumerate}


\section*{Data Availability Statement}

The datasets analyzed during the current study are available in the following repositories:
\begin{itemize}
    \item The Cardiovascular Disease Dataset is publicly available on Kaggle at \url{https://www.kaggle.com/datasets/sulianova/cardiovascular-disease-dataset}.
    \item The MIMIC-III Clinical Database, which was used to derive the Sepsis Prediction, Hospital Length of Stay, and Mortality Prediction datasets, is available on PhysioNet at \url{https://physionet.org/content/mimiciii/1.4/}. Access requires completion of a data use agreement and the CITI "Data or Specimens Only Research" course.
    \item The Pan-Cancer Atlas (PanCan) dataset is publicly available from the GDC Data Portal at \url{https://gdc.cancer.gov/about-data/publications/pancanatlas}.
    \item The Breast Cancer (classification and survival) and Lung Cancer datasets are included within the scikit-learn and sksurv Python libraries, respectively, and are publicly available.
\end{itemize}